# DYNAMICAL PREDICTIVE MODELLING OF CARDIOVASCULAR DISEASE PROGRESSION POST-MYOCARDIAL INFARCTION VIA ECG PRE-TRAINED ARTIFICIAL INTELLIGENCE MODEL

**Riccardo Cavarra[1], Lupo Lovatelli[1], Shaheim Ogbomo-Harmitt[1], Shahid Aziz[2], Adelaide De Vecchi[1], Andrew King[1] and Oleg Aslanidi[1]**
[1]King's College London, St Thomas' Hospital, London, UK, {riccardo.cavarra, lupo.lovatelli, shaheim.ogbomo-harmitt, adelaide.de_vecchi, andrew.king, oleg.aslanidi}@kcl.ac.uk
[2]North Bristol NHS Trust, Bristol, UK, {shahid.Aziz}@nbt.nhs.uk

**SUMMARY**

Myocardial infarction (MI) is a leading cause of death, and its adverse outcomes are urgent to predict. Yet ECG-based prognostic models underperform because deep learning requires large, labelled datasets, which are scarce in medicine. Foundation models can learn from unlabelled ECGs via self-supervision, but medically relevant training strategies remain underexplored. We propose a pre-trained artificial intelligence model that combines patient-specific temporal information using contrastive learning with supervised multitask heads, then fine-tunes on post-MI outcome prediction. The proposed model outperformed a model trained from scratch (0.794 vs 0.608 AUC) showing that clinically structured ECG modelling improves classification in limited data regimes.

## 1 INTRODUCTION

Cardiovascular diseases (CVD) are the leading cause of death and long-term morbidity, affecting the quality of life of millions of people and posing a significant burden on public health. Among these, acute myocardial infarction (MI) is particularly concerning due to high mortality rates and potential for lasting disability. Early identification of high-risk individuals is therefore essential for early intervention and management of CVD [1]. The electrocardiogram (ECG) is a low-cost, non-invasive measurement of cardiac electrophysiology and provides insights into the severity and progression of MI. Despite this, its use is mostly limited to the early stratification of potential MI cases.

In clinical practice the ECG has no role in patient management following MI [2]. Understanding the complex information carried by ECG signals can help us identify patients at high risk of death and comorbidity and ultimately improve care. Artificial intelligence (AI) and particularly deep learning (DL) have been extensively used to model complex data relying on human-made labels in a supervised fashion [3]. These, however, are underexplored in the medical field due to the lack of large, diverse and annotated data. In such data regimes, DL strategies fail to extract clinically relevant information from medical data and rely on shortcut to make predictions, leading to underperformance and poor generalisation on unseen data. This warrants the need for representation learning strategies that exploit the scale of unlabelled ECG archives while retaining clinically meaningful structure.

Foundation models can leverage contrastive learning to model medical data by learning reusable, medically relevant representations [4]. They do so by maximising the similarity between related signals while separating unrelated ones, where related signals are usually different augmented views of the same instance [5]. Despite promising results in other fields, these general approaches do not incorporate patient-level temporal context. In post-MI cohorts, clinically relevant ECG variation reflects disease progression and a representation that is sensitive to disease-relevant change is more suitable for downstream prognosis. Hence, we argue that patient-level temporal context is paramount in ECG modelling and we reframe contrastive learning as a dynamical modelling problem.

The aim of this paper is to develop an ECG AI model that learns patient-specific, dynamics-aware representations by combining contrastive learning with supervised multitask objectives. By defining contrastive positives as ECGs recorded from the same patient within a clinically meaningful time window, the model is encouraged to capture disease-relevant change over time while remaining invariant to acquisition noise and nuisance variability. We then fine-tune this large pre-trained model on smaller post-MI cohorts to predict two adverse outcomes (mortality and heart failure) demonstrating that clinically structured ECG representation learning improves downstream prognostic performance in limited data regimes.

## 2 METHODOLOGY

Patient data were obtained from the Medical Information Mart for Intensive Care (MIMIC-IV) database, which contains electronic health records and waveform data collected between 2008 and 2019 at the Beth Israel Deaconess Medical Centre in Boston [6]. From this resource, we extracted 800,035 12-lead ECG recordings from 161,352 unique patients, which were used to pre-train the AI model. For the downstream tasks we considered two separate subsets of patients diagnosed with MI, identified through standardised diagnoses reports that use the international classification of diseases (ICD-10: I21, ICD-9: 410). For these tasks, a single ECG signal recorded after MI diagnosis was used. The first subset included 8499 MI patients, of whom 3,097 died during the follow up, and was used to predict mortality determined using in-hospital and post-discharge mortality information. The second subset was used to predict heart failure instances following MI and it included 852 patients, of whom 552 developed heart failure (ICD-9: 4280, 42820, 42821, 42822, 42823). All ECG recordings were 10 seconds long and sampled at 500 Hz. Preprocessing involved low-pass (150 Hz) and high-pass filtering (0.5 Hz) and z-score normalisation.

### 2.1 Model pre-training

The aim of the proposed model is to model ECG signals preserving patient-specific temporal context using a combination of contrastive learning and multitask supervised learning. The outline of the model's architecture is shown in Figure 1. Pairs of ECG signals are sampled from the dataset and transformed using stochastic data augmentation creating distorted but clinically meaningful representations of the same signal. These augmentations prevent shortcut solutions and promote invariances consistent with ECG acquisition and variability. The augmentation family includes additive Gaussian noise, global amplitude scaling, small circular time shifts, time masking, lead dropout, and baseline wander. Two different augmentation transforms $(t, t')$ are applied to the ECGs in the pair $(\tilde{x}_i, \tilde{x}_j)$ which are then fed into a 1D convolutional encoder (ResNet1D) [7]. This encoder uses strided convolution, followed by residual blocks and global average pooling to extract information from the ECG signal, producing a fixed embedding which is then passed to a projection head with two dense layers and batch normalisation. The output of this projection head is a condensed representation of the ECG signals, $(z_i, z_j)$. The learning process is governed by the contrastive loss, which is computed on these embeddings and encourages the encoder to extract clinically relevant information from the ECG by maximising the similarity between positive pairs of ECGs. As shown in equation 1, a pair of ECG signals is considered positive if both signals come from the same patient $s$ and were recorded during the same temporal window of $T_w = 60$ days. Pairs of ECGs coming from different patients or recorded far away in time are treated as negative pairs.

$$\mathcal{P} = \{(i,j): s_i = s_j, \ | t_i - t_j | \leq T_w\} \qquad \text{Eq. 1}$$

During training, each minibatch samples $B$ such pairs and expands them into $N = 2B$ examples in the fixed order $(i_0, j_0, i_1, j_1, \dots, i_k, j_k, \dots)$, making the positive index $p(k) = k \oplus 1$ (XOR with 1). Thus, for every $k$ ECG signal, the model learns to find its positive partner $p(k)$. ECG signal similarity is assessed across all positive pairs in the batch using the normalised, temperature scaled cross entropy loss (NT-Xent) [5], also referred to as self-supervised learning loss ($\mathcal{L}_{\text{ssl}}$) shown in equation 2.

$$\mathcal{L}_{ssl} = -\frac{1}{N}\sum_{k=1}^{N} log \frac{exp\left(\frac{\tilde{z}_k^{\top}\tilde{z}_{p(k)}}{\tau}\right)}{\sum_{\ell \neq k} exp\left(\frac{\tilde{z}_k^{\top}\tilde{z}_{\ell}}{\tau}\right)}, \quad \tilde{z} = \frac{z}{\|z\|_2} \qquad \text{Eq. 2}$$

Here $\tau = 0.2$ is a temperature parameter. This contrastive loss maximises the cosine similarity between positive pairs of ECGs while increasing the separation between negative pairs. By doing this, we condition the training paradigm to learn patient-specific temporal context that helps the model understand disease progression in the context of MI while also learning invariance to ECG noise thanks to the stochastic augmentations.
In addition to self-supervision, we impose clinically meaningful structure by training linear task heads on the encoder embedding $z$. We define two linear task heads to perform classification and regression. They respectively learn to classify ECGs based on rhythm disturbances (atrial fibrillation and ST segment elevation) and measure the median duration of the ECG's waveforms (P-wave, QRS-complex, T-Wave, RR interval). Regression uses a masked Huber loss $(\mathcal{L}_{reg})$ to handle missing targets while classification uses binary cross entropy loss $(\mathcal{L}_{cls})$. To combine heterogeneous tasks with different scales and noise levels, we use Kendall-style homoscedastic uncertainty weighting [8] which scales each objective based on their task-dependent uncertainty. Let $s_k = log\ \sigma_k^2$ be a learned log-variance parameter for regression target $k$, and $r_m = log\ \sigma_m^2$ for classification target m. The supervised objective $(\mathcal{L}_{sl})$ is given by equation 3.

$$\mathcal{L}_{sl} = \sum_k \left(exp\ (-s_k)\ \mathcal{L}_k^{reg} + s_k\right) + \sum_m \left(exp\ (-r_m)\ \mathcal{L}_m^{cls} + r_m\right) \qquad \text{Eq. 3}$$

These supervised tasks act as a gentle regulariser that guides the model and help its features remain clinically relevant. The model is pretrained by jointly optimising $\mathcal{L}_{\text{ssl}}$ and $\mathcal{L}_{\text{sup}}$ as shown in equation 4.

$$\mathcal{L} = \lambda_{ssl}\mathcal{L}_{ssl} + \lambda_{sl}\mathcal{L}_{sl} \qquad \text{Eq. 4}$$

with $\lambda_{ssl} = 10$ and $\lambda_{sup} = 0.2$. Training uses mixed precision on an RTX 4090 GPU and AdamW optimisation. Early stopping was used to interrupt training when the model started overfitting. We monitor representation quality via in-batch retrieval accuracy (whether the nearest neighbouring ECG signal is indeed the paired positive) and by tracking positive/negative cosine similarities.

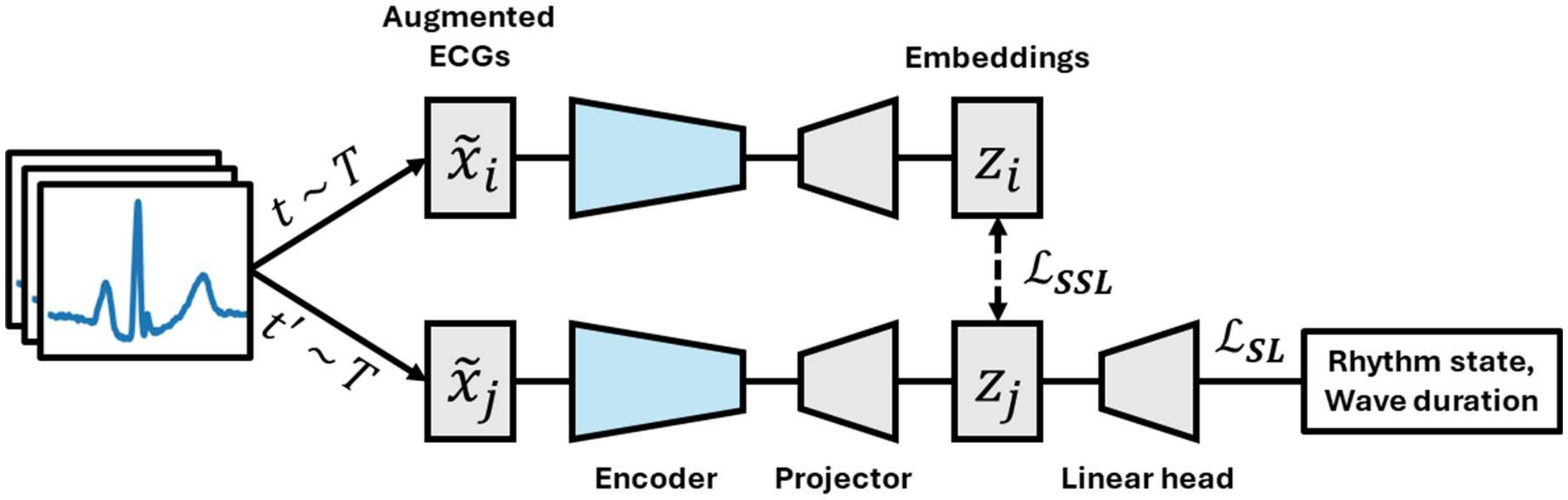

Figure 1: Architecture outline of the proposed large, pre-trained AI model. Two augmentation operators $(t, t')$ of the same family $T$ are applied to samples ECG signals to obtain contrastive views $(\tilde{x}_i, \tilde{x}_j)$. The encoder (ResNet1D) and the projector learn to embed these signals $(z_i, z_j)$ to maximise the cosine similarity between ECGs sampled from the same patient at different times in a self-supervised fashion. Separate linear heads learn to predict the rhythm state and wave duration of the ECGs in a supervised fashion to regularize the training.

### 2.2 Adverse outcome classification

Following pretraining using 800,035 ECGs, the model is fine-tuned to classify mortality and heart failure instances post-MI. The ResNet1D encoder of the pre-trained model and its weights are transferred to an ECG classifier. The pretrained encoder extracts meaningful features from the ECG signal while a network of dense layers is trained to classify signals using these features. To assess the performance of the model we compare these results to an identical network trained from scratch to perform the same tasks.

## 3 RESULTS AND CONCLUSIONS

We evaluate the transfer learning capabilities of our proposed model by comparing it with a ResNet1D trained from scratch. Table 1 illustrates the performance of the models on two downstream tasks: mortality and heart failure prediction post-MI using single ECG instances. The performance of the models was assessed using the area under the receiver operating characteristic curve (AUC). Results show that pretraining the ResNet1D encoder on a clinically relevant contrastive task improves the model's understanding of ECG structure and morphology and helps it achieve better classification accuracy in downstream tasks. Pretraining improved the AUC by 19% in mortality prediction and by 18% in heart failure classification when testing the model on unseen data.

Table 1: Comparison of transfer learning performance of the pre-trained AI model with a baseline ResNet1D trained from scratch. Model performance is measured using AUC and evaluated on two classification tasks, mortality and heart failure prediction following MI. AUC: area under the receiver operating characteristic curve.

| Task | AUC | |
|---|---|---|
| | Training from scratch | Pre-trained AI model |
| Mortality prediction | 0.608 | **0.794** |
| Heart failure prediction | 0.602 | **0.781** |

In conclusion, this study investigated a novel pipeline to model the ECG signal using a large pre-trained AI model to improve the prediction of adverse outcomes post-MI, mortality and heart failure. Our results across two different classification tasks consistently show that effective ECG modelling using self-supervised learning can improve long term outcome prediction post-MI and serves as a proof-of-concept study for the use of clinically smart pre-trained AI models in low data regimes like CVD outcome prediction.